\documentclass[letterpaper,10pt,conference]{ieeeconf}
\usepackage[T1]{fontenc}
\usepackage[utf8]{inputenc}
\usepackage{amsmath,amssymb,mathtools}
\usepackage{booktabs}
\usepackage{array}
\usepackage{algorithm}
\usepackage[noend]{algpseudocode}
\usepackage{graphicx}
\usepackage{tikz}
\usepackage{url}
\usepackage{cite}
\usepackage[hidelinks]{hyperref}
\hypersetup{pdfauthor={Vincent Katter and Marnix Suilen and Pieter De Clercq and Walter Daems and Jan Steckel and Guillermo A. P\'{e}rez},pdftitle={Synchronize or Reveal: Planning for Ambiguous LiDAR Localization}}
\newtheorem{assumption}{Assumption}
\newtheorem{definition}{Definition}
\newtheorem{problem}{Problem}
\newtheorem{proposition}{Proposition}
\newcommand{\formalbodyfont}{\normalfont}
\AddToHook{env/proposition/begin}{\renewcommand{\formalbodyfont}{\normalfont\itshape}}
\makeatletter
\def\@begintheorem#1#2{%
  \@IEEEtmpitemindent\itemindent
  \topsep 4pt plus 1pt minus 1pt\formalbodyfont\trivlist
  \item[\hskip\labelsep{\normalfont\bfseries #1\ #2.}]%
  \itemindent\@IEEEtmpitemindent}
\def\@opargbegintheorem#1#2#3{%
  \@IEEEtmpitemindent\itemindent
  \topsep 4pt plus 1pt minus 1pt\formalbodyfont\trivlist
  \item[\hskip\labelsep{\normalfont\bfseries #1\ #2\ (#3).}]%
  \itemindent\@IEEEtmpitemindent}
\def\@endtheorem{\endtrivlist}
\makeatother
\newcommand{\Model}{\mathcal M}
\newcommand{\Fam}{\mathcal B}
\newcommand{\Obs}{\Omega}
\newcommand{\allowed}{A_{\mathrm c}}
\newcommand{\outcome}[1]{\textnormal{\texttt{#1}}}
\title{\LARGE\bf Synchronize or Reveal: Planning for Ambiguous LiDAR Localization}
\author{
\authorblockN{Vincent Katter$^{1,2}$,
Marnix Suilen$^{3}$,
Pieter De Clercq$^{2,4}$,
Walter Daems$^{1,2}$,
Jan Steckel$^{1,2}$,
Guillermo A. P\'{e}rez$^{3}$}
\authorblockA{$^{1}$Cosys-Lab, Faculty of Applied Engineering, University of Antwerp, Antwerp, Belgium}
\authorblockA{$^{2}$Flanders Make Strategic Research Centre, Lommel, Belgium}
\authorblockA{$^{3}$AnSyMo, Department of Computer Science, Faculty of Science, University of Antwerp, Antwerp, Belgium}
\authorblockA{$^{4}$Department of Computer Science, KU Leuven, Leuven, Belgium\\
Corresponding author: vincent.katter@uantwerpen.be}
}
\begin{document}
\AddToHookNext{shipout/foreground}{\put(0.5\paperwidth,-0.45in){\makebox[0pt]{\parbox{0.8\paperwidth}{\centering\footnotesize This work has been submitted to the IEEE for possible publication. Copyright may be transferred without notice, after which this version may no longer be accessible.}}}}
\maketitle
\thispagestyle{empty}
\pagestyle{empty}
\suppressfloats[t] %

\begin{abstract}
Repeated warehouse geometry can leave a robot with several plausible positions and headings.
Actions can resolve this ambiguity by bringing the alternatives into one configuration or by producing different observations.
In this work, we search a deterministic model of zones and headings for action sequences that result in at most a requested number of configurations after any possible observation sequence.
A monitor checks that each localization update preserves this requested limit for the rest of the sequence.
The limit also holds during execution if the robot's movements and sensor observations match the model and its true configuration remains among the possibilities.
We implemented our approach in a Gazebo simulator.
Our experiments show that planned movements and observations of passages, junctions, and blocked paths allow the robot to resolve its uncertainty and reach its destination.
\end{abstract}

\section{Introduction}
\label{sec:introduction}
Partial or noisy sensor observations can leave a robot uncertain about its location.
For example, a LiDAR scan may match several warehouse aisles with similar geometry.
Its possible poses may also face opposite directions.
A forward command then takes one pose hypothesis toward an opening and another toward a wall.
Choosing either pose prematurely can send the robot the wrong way.
A \emph{support} represents the robot's possible configurations.
Multi-hypothesis localization can supply such alternatives to a navigation system~\cite{Garcia2023MultiHypothesis}.

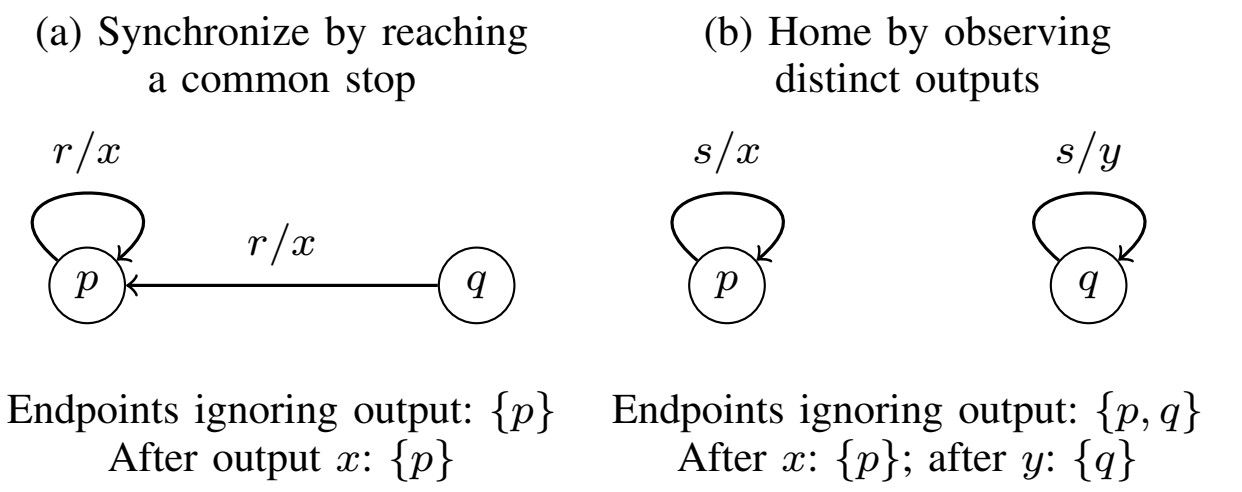
\begin{figure}[t]
\centering
\begin{tikzpicture}[x=0.92mm,y=1mm,every node/.style={font=\footnotesize},
  state/.style={circle,draw,minimum size=5mm,inner sep=0pt},
  transition/.style={->,semithick}]
\node[align=center,font=\footnotesize] at (19,10)
  {(a) Synchronize by reaching\\a common stop};
\node[align=center,font=\footnotesize] at (64,10)
  {(b) Home by observing\\distinct outputs};
\node[state] (rp) at (5,-5) {$p$};
\node[state] (rq) at (33,-5) {$q$};
\draw[transition] (rq) -- (rp) node[midway,above] {$r/x$};
\draw[transition] (rp) to[out=140,in=40,looseness=6] node[above] {$r/x$} (rp);
\node[state] (sp) at (51,-5) {$p$};
\node[state] (sq) at (77,-5) {$q$};
\draw[transition] (sp) to[out=140,in=40,looseness=6] node[above] {$s/x$} (sp);
\draw[transition] (sq) to[out=140,in=40,looseness=6] node[above] {$s/y$} (sq);
\node[align=center] at (19,-15) {Endpoints ignoring output: $\{p\}$\\After output $x$: $\{p\}$};
\node[align=center] at (64,-15) {Endpoints ignoring output: $\{p,q\}$\\After $x$: $\{p\}$; after $y$: $\{q\}$};
\end{tikzpicture}
\caption{Two ways to identify the final configuration.
Both panels start from $H=\{p,q\}$; arrows show action/output.
Action $r$ sends both possibilities to $p$.
Action $s$ preserves their positions, and its output distinguishes $p$ from $q$.
A self-loop leaves the configuration unchanged.}
\label{fig:mechanisms}
\end{figure}

The robot has two ways to reduce its number of hypotheses.
It can execute a sequence that brings every alternative into one zone and heading class.
This process is \emph{synchronization}: the final abstract configuration is known even if the observations do not distinguish the alternatives.
(For zone-only goals, a final action can map all goal-zone alternatives to one heading class, making heading a modeling technicality.) 
It can also move so that different alternatives produce different sensor responses.
A \emph{homing} sequence ensures that those responses identify the final abstract configuration, which can depend on the response received.
Both mechanisms may act within the same sequence.
Figure~\ref{fig:mechanisms} illustrates them in a complete two-state example.

\textbf{Our contributions.}
We develop ambiguity-resolution planning for a supervisor that coordinates a motion controller and a localizer.
A finite deterministic model predicts the effects and outputs of the controller's actions.
A successful planning call returns a \emph{fixed word}, a finite action sequence shared by all possible observation histories.
Replaying the word from each possible start bounds the number of final configurations for every output history.
Our algorithm retains a family of posterior supports, one for each possible output history, and seeks a common word that reduces every member to the requested size.
It operates in two phases: First, it relies on precomputed pair-resolving sequences. Then, it falls back to bounded exact search.
The planner distinguishes a verified plan, exhaustive failure, and an exhausted budget.
We give sufficient conditions for localization updates to preserve the plan's ambiguity bound, and use them to monitor execution and replan.
Experiments on a Gazebo-simulated warehouse show how motion, coarse outputs, and localization reduce ambiguity.
Furthermore, we compare fixed and adaptive plans, and test the execution assumptions.
While we focus on LiDAR mainly, we expect the architecture to extend to sonar or camera localization using sensor-specific classifiers and action/output models.

\textbf{Related work.}
Deterministic partially observable Markov decision processes (det-POMDPs) and their generalizations capture both convergence under actions and uncertainty reduction through observations~\cite{Bonet2009DetPOMDP,DBLP:journals/corr/abs-2602-07473}.
Their adaptive policies (which are computationally expensive to obtain) choose actions from observation histories, whereas our fixed words use one action sequence for every history.
Robotics approaches instead exploit local observability for task decomposition~\cite{Merlin2024LocalObservability}, guide continuous belief search with simplified models~\cite{Ho2023SiMBA}, reduce data-association hypotheses with bounded error~\cite{Shienman2022D2A}, choose informative motions or sensing actions~\cite{Fox1998ActiveMarkov,Tzes2022Informative}, or learn macro-actions~\cite{Liang2023RMAG}.
Our objective is a worst-case support-size bound from one fixed sequence.
Our techniques draw directly from automata testing: synchronization and homing supply the objectives~\cite{Eppstein1990,LeeYannakakis1994,Volkov2008Cerny}, while pair-based heuristics concatenate resolving sequences~\cite{Cirisci2019Homing}.

\section{Model and Problem}
\label{sec:model}
Consider a robot whose LiDAR scan matches several warehouse aisles and headings.
The robot can drive forward or turn, then use the next scan to discard alternatives.
It can also move so that several alternatives reach the same place.
We seek action sequences that reduce this uncertainty and allow the robot to continue toward its destination.

\subsection{Configurations, Actions, and Outputs}
The warehouse map groups locations into zones such as aisles and intersections.
Forward motion also depends on the robot's heading: facing the other way changes which zone it approaches.
We therefore represent an \emph{abstract configuration} by $q=(z,h)$, where $z$ is a zone and $h$ is a heading class.
The warehouse model assigns headings to the nearest of north, east, south, and west.
Let $Q$ be the finite set of abstract configurations.
The LiDAR \emph{localizer} combines scans with motion information to maintain pose hypotheses; mapping their reported centroids to $Q$ gives a set $L\subseteq Q$.
The supervisor maintains a nonempty \emph{support} $H\subseteq Q$, initialized from a localizer report or supplied separately.
A singleton support identifies one zone and heading class.

An action in $A$ denotes a controller action, e.g., driving forward to the next zone or turning left by $90^\circ$.
After the action, the robot records an \emph{(output) observation} in $\Obs$.
For a LiDAR scan $s$, we have a \emph{coarse (or abstract) classifier} that returns $c(s)\in\Obs$: junction when at least three of four directions around the robot appear open, passage otherwise.
After any action and subsequent LiDAR scan $s$, the observation $c(s)$ is reported; unless forward motion is obstructed, then \texttt{blocked} is reported instead.
To model how each hypothesis moves and what the robot observes, we use a deterministic Mealy machine~\cite{LeeYannakakis1994}, $\Model=(Q,A,\Obs,\delta,\eta)$, with transition and output functions
\begin{equation}
\delta \colon Q\times A\to Q,\quad \eta \colon Q\times A\to\Obs.
\label{eq:model}
\end{equation}
For configuration $q$ and action $a$, $\delta(q,a)$ is the successor configuration and $\eta(q,a)$ is the output.
For the warehouse, the model derives junction and passage outputs from the successor zone's neighbors as described above.
We assume both functions are total: each configuration-action pair has one successor and one output.
A safely blocked motion can be represented by a self-loop with output \texttt{blocked}.

\subsection{Words as Plans}
Each planning request specifies a nonempty set of permitted actions $\allowed\subseteq A$.
A \emph{word} $w=a_1\cdots a_m\in\allowed^*$ is a finite sequence of actions, or a nonreactive plan.
A run of the Mealy machine on such a word, starting at $q_0=q$, satisfies
\begin{equation}
\begin{aligned}
q_i&=\delta(q_{i-1},a_i),\quad i=1,\ldots,m,\\
\operatorname{obs}(q,w)&=\eta(q_0,a_1)\cdots\eta(q_{m-1},a_m).
\end{aligned}
\label{eq:trace}
\end{equation}
Write $\delta(q,w)=q_m$; the \emph{output trace} $\operatorname{obs}(q,w)$ belongs to $\Obs^{|w|}$.
We use $\varepsilon$ for both the empty action sequence and the empty output sequence, with $\delta(q,\varepsilon)=q$ and $\operatorname{obs}(q,\varepsilon)=\varepsilon$.

The \emph{motion image} $\delta(H,w)$ collects all possible endpoints.
For an output trace $o\in\Obs^{|w|}$, we retain endpoints reached by hypotheses that predict it:
\begin{align}
\delta(H,w)&=\{\delta(q,w) : q\in H\},\label{eq:image}\\
H_{w,o}&=\{\delta(q,w):q\in H,\ \operatorname{obs}(q,w)=o\}.
\label{eq:posterior}
\end{align}
The second set is the \emph{posterior support}.
For example, a junction output removes hypotheses that predict a passage at that step.
We measure the plan's residual ambiguity by the largest posterior support over its possible traces:
\begin{equation}
U_H(w)=\max_{o\text{ attainable from }H}|H_{w,o}|.
\label{eq:ambiguity}
\end{equation}
\begin{definition}[Support objectives]
\label{def:objectives}
Let $k\geq1$ be an integer.
A word $w\in\allowed^*$ \emph{synchronizes} $H$ if $|\delta(H,w)|=1$.
It is \emph{$k$-disambiguating} if $U_H(w)\leq k$, and \emph{strictly disambiguating} if $U_H(w)<|H|$.
It \emph{homes} $H$ if $U_H(w)=1$.
\end{definition}

Thus homing identifies a final configuration\footnote{In contrast, for the expert reader familiar with the literature,
a \emph{distinguishing} word identifies the initial configuration~\cite{LeeYannakakis1994}.} for every attainable trace.
The supports satisfy
\begin{equation}
1\leq U_H(w)\leq|\delta(H,w)|\leq|H|.
\label{eq:bounds}
\end{equation}
For a nonsingleton support, synchronization implies homing, which implies strict disambiguation.
Both convergence and informative outputs can decrease $U_H(w)$.
The strict relation $U_H(w)<|\delta(H,w)|$ holds when observations improve the worst-case bound beyond motion alone for that word.

In Figure~\ref{fig:mechanisms}(a), $r$ synchronizes and homes. %
In panel (b), $s$ homes without synchronization.
If $s$ is removed, the two states have identical output behavior on all actions (namely, just $r$). 
However, $r$ still synchronizes them.

\begin{problem}[Fixed-word ambiguity resolution]
\label{prob:resolution}
Given $\Model$, a nonempty support $H\subseteq Q$, a nonempty permitted alphabet $\allowed\subseteq A$, and an integer $1\leq k\leq|H|$, find a word $w\in\allowed^*$ with $U_H(w)\leq k$, or establish that none exists.
For synchronization, use the objective $|\delta(H,w)|=1$.
\end{problem}

\textbf{From configurations to physical poses.}
Actions are controller-level commands with completion conditions, such as driving to the next zone or turning by $90^\circ$.
For each $q$, $\gamma(q)$ contains permitted initial poses and controller endpoints; the modeled action effect and output must agree across them.
\begin{assumption}[Execution abstraction]
\label{ass:execution}
For every $q\in Q$, every pose $x\in\gamma(q)$, and every permitted action $a$ used by the plan, execution terminates safely in $\gamma(\delta(q,a))$.
On completion, the sampling protocol records exactly the output $\eta(q,a)$.
Let \(q_0 \in H\) and \(x_0 \in \gamma(q_0)\) be the true initial configuration and pose. After each completed action prefix
\(w_i\), the support \(R_i\) accepted following a localizer refinement contains the true successor configuration: \(\delta(q_0,w_i)\in R_i\).
\end{assumption}

The successor condition keeps every permitted endpoint inside the represented sets, including executions from alternative hypotheses.
Section~\ref{sec:validity} evaluates observation agreement on the warehouse regions.

\subsection{Reaching a Goal}
The warehouse robot's target is a zone.
Let $T\subseteq Q$ contain its accepting configurations, including allowed headings.
\emph{Goal containment} requires $\delta(H,w)\subseteq T$; \emph{synchronous arrival} additionally requires $|\delta(H,w)|=1$.
A nonsingleton support already contained in $T$ meets the first objective with $\varepsilon$, but may not meet the second.

After executing a homing word, the output trace identifies the robot's final configuration.
A path to the goal can subsequently be selected from the configuration, if one exists.
More formally, let \(w\) be a homing word for \(H\). By definition, each attainable output trace \(o\) identifies a unique final configuration \(q_o\).
If each such \(q_o\) has a path $v_o$ into $T$, then $(w,\{v_o\}_o)$ is a \emph{contingent policy} with a choice after $w$.
Every branch of this policy ends in $T$, with worst-case action count $|w|+\max_o|v_o|$.
We note that a zone-only target can also be encoded as synchronous arrival.\footnote{For a heading-independent goal zone $z_g$, add an action $a_g$ and choose $h_g$ so that $\delta((z_g,h),a_g)=(z_g,h_g)$ for every $h$ admitted by $T$. Then $\delta(H,w)\subseteq T$ implies $|\delta(H,wa_g)|=1$, if $a_g$ satisfies Assumption~\ref{ass:execution}.}

\section{Planning to Resolve Ambiguity}
\label{sec:planning}
A fixed word must meet the requested ambiguity bound on every output branch using the same actions.
The planner first tries words that resolve two hypotheses and checks their effect on the whole support.
Such a word, called a \emph{pair witness}, either brings the two hypotheses together or produces different output traces from them.
If no candidate succeeds, exact search restarts from $H$.
\subsection{Candidate Words from Pairs}
Preprocessing stores pair witnesses for use in later planning requests.
The pair graph contains unordered pairs $\{p,q\}$, including diagonal pairs with $p=q$, and edges
\begin{equation}
\{p,q\}\xrightarrow{a}\{\delta(p,a),\delta(q,a)\}.
\label{eq:pair}
\end{equation}
Reverse breadth-first search (BFS) from the diagonals constructs a merging-witness table, recording a shortest merging witness for each \emph{mergeable} pair~\cite{Eppstein1990,Volkov2008Cerny}.
Observation can also resolve a pair when the two next outputs differ.
\begin{definition}[Pair resolution]
\label{def:pair-resolution}
Action $a$ resolves $\{p,q\}$ if
\begin{equation}
\delta(p,a)=\delta(q,a)\quad\text{or}\quad\eta(p,a)\ne\eta(q,a).
\label{eq:resolve}
\end{equation}
\end{definition}

A resolution-witness table uses Definition~\ref{def:pair-resolution} to terminate at either a merge or different outputs.
The pair graph has $|Q|(|Q|+1)/2$ nodes.
Its construction and reverse searches require $O(|A||Q|^2)$ time and space, storing first-action and successor-pair pointers.
If $a$ takes a pair to $\{u,v\}$ with equal outputs and $b$ emits different outputs from $u,v$, reverse search records the two-action resolving witness $ab$.

For synchronization, a greedy routine evaluates merging witnesses on the entire active support.
It chooses the smallest resulting support, breaking ties by word length and pair index, and repeats after updating the support.
Restricted-action synchronization obtains its pair witnesses within $\allowed$.
If this phase stalls, exact forward BFS restarts from the original $H$ and seeks a singleton motion image.
The fallback keeps transitions that preserve current-support size---since a necessary move may not immediately merge any states.
The restart lets exact search consider alternatives to the greedy choices.
Our two-phase heuristic search is warranted since deciding whether a support has a synchronizing word is PSPACE-complete~\cite{Vorel2016SubsetSync}.

For disambiguation, the planner retrieves one resolution witness for each pair in the original $H$ and replays the word from every state in $H$.
It ranks candidates by increasing $U_H(w)$, increasing length, decreasing number of distinct output traces, and finally pair index.
The selected candidate must strictly reduce $U_H$ and meet the requested bound. %
If no candidate satisfies the requested ambiguity bound, the planner resorts to exact search.

The stored table may cover a larger alphabet than the deployment permits.
The planner first selects its best stored candidate, then rejects it if any letter is outside $\allowed$, which leads directly to exact search from $H$.

\subsection{Exact Search over Possible Posterior Supports}
The same action sequence can produce different output traces, depending on the initial configuration.
Each trace defines a posterior support \(H_{w,o}\).
To construct a fixed word, the search tracks all possible posterior supports and extends them with the same action.
For a support $S$, define its successor conditioned on one output by
\begin{equation}
S[a,o]=\{\delta(q,a):q\in S,\ \eta(q,a)=o\}.
\label{eq:branch}
\end{equation}
A search state is a family $\Fam$ of nonempty posterior supports, initially $\{H\}$.
A common action has successor
\begin{equation}
F_a(\Fam)=\{S[a,o]:S\in\Fam,\ o\in\Obs,\ S[a,o]\ne\varnothing\}.
\label{eq:family}
\end{equation}
These supports can overlap in final configurations.
For example, consider a word $ab$ with the following traces and endpoints from five starting configurations:
\begin{center}
\begin{tabular}{@{}ccc@{}}
\toprule
Initial configuration & Complete trace & Final configuration\\
\midrule
$h_1,h_2$ & $xz$ & $e_1$\\
$h_3$ & $xz$ & $e_2$\\
$h_4$ & $yz$ & $e_2$\\
$h_5$ & $yz$ & $e_3$\\
\bottomrule
\end{tabular}
\end{center}
Motion reduces five alternatives to three endpoints, $\delta(H,ab)=\{e_1,e_2,e_3\}$.
Conditioning on the traces gives $\Fam=\{\{e_1,e_2\},\{e_2,e_3\}\}$ and $U_H(ab)=2$.
Endpoint $e_2$ is compatible with both past traces.
If a common action $c$ takes $e_1,e_2,e_3$ to $t$ with one output, both supports become $\{t\}$.

Identical supports have identical futures under any common suffix, so their multiplicity and history names do not matter. %
We implement supports as bitsets, remove duplicates, and sort them to compare families during search.

\begin{algorithm}[t]
\caption{Fixed-word disambiguation}
\label{alg:disambiguation}
\begin{algorithmic}[1]
\Require Model $\Model$, nonempty $H$, alphabet $\allowed$, integers $1\leq k\leq|H|$ and $B\geq1$
\If{$|H|\leq k$}
  \State \Return \outcome{ALREADY\_SATISFIED} with $\varepsilon$
\EndIf
\State $u\gets\Call{BestPairProposal}{H}$ and $e\gets0$
\If{$u$ exists, $u\in\allowed^*$, and $U_H(u)<|H|$}
  \State $e\gets1$
  \If{$U_H(u)\leq k$}
    \State \Return \Call{Verify}{$H,u,k$}
  \EndIf
\EndIf
\State queue $\{H\}$, mark it visited, and save word $\varepsilon$
\While{queue nonempty and $e<B$}
  \State pop family $\Fam$ with saved word $w$; $e\gets e+1$
  \ForAll{$a\in\allowed$}
    \State $\Fam'\gets\Call{Canonicalize}{F_a(\Fam)}$
    \If{$\Fam'$ has not been visited}
      \State mark $\Fam'$ visited and save word $wa$
      \If{$\max_{S\in\Fam'}|S|\leq k$}
        \State \Return \Call{Verify}{$H,wa,k$}
      \EndIf
      \State enqueue $\Fam'$
    \EndIf
  \EndFor
\EndWhile
\If{queue empty}
  \State \Return \outcome{NO\_PLAN}
\Else
  \State \Return \outcome{RESOURCE\_BOUND}
\EndIf
\end{algorithmic}
\end{algorithm}

Algorithm~\ref{alg:disambiguation} gives the bounded search for Problem~\ref{prob:resolution}'s $k$-disambiguating objective.
\textsc{Verify} computes Equation~\eqref{eq:posterior} for every attainable trace and returns \outcome{PLAN} only if the requested bound holds.
BFS chooses one common action at each level.
Allowing a separate next action for each family member would synthesize an adaptive policy.

\subsection{Results, Budgets, and Correctness}
The planner returns a word meeting the requested objective with status \outcome{PLAN}, or the empty word with status \outcome{ALREADY\_SATISFIED} when $H$ already meets it.
It reports \outcome{NO\_PLAN} only after exhaustive search of the relevant finite support graph.
It reports \outcome{RESOURCE\_BOUND} when work remains after the expansion budget is exhausted.

The two phases use a shared budget \(B\).
An improving disambiguation proposal consumes one expansion even when it misses $k$ and is discarded.
Subsequent expansions count dequeued families.
Synchronization similarly counts accepted greedy merges and fallback support expansions.
The expansion counts exclude witness retrieval, candidate replay, pair searches, and preprocessing.
Heuristic words need not be shortest.
A successful BFS fallback uses the fewest symbolic actions for its fixed-word objective.

\begin{proposition}[Word verification \& execution]
\label{prop:word}
On a finite deterministic model, acceptance by the verifier implies that $|\delta(H,w)|=1$ for a synchronization request, or $U_H(w)\leq k$ for a $k$-disambiguation request.
Under Assumption~\ref{ass:execution}, the same bound on final abstract configurations holds when the robot executes the word.
Exhaustive family BFS proves absence of a fixed $k$-disambiguating word over $\allowed$ when it returns \outcome{NO\_PLAN}.
\end{proposition}
\begin{proof}
The verifier simulates $w$ from each configuration in $H$.
For each output trace $o$, it collects the distinct final configurations of runs producing $o$.
This gives the posterior support $H_{w,o}$ in Equation~\eqref{eq:posterior}, whose size the verifier counts.
Under the execution assumption, induction over action completions maps each concrete execution to that same abstract run and trace.
Writing $\Fam_H(w)=\{H_{w,o}:o\text{ attainable from }H\}$, we have $\Fam_H(\varepsilon)=\{H\}$ and $\Fam_H(wa)=F_a(\Fam_H(w))$ by Equation~\eqref{eq:family}.
Duplicate supports have identical successors and cannot affect the maximum cardinality after any suffix.
BFS thus visits every relevant family if its frontier is exhausted, and a successful family exists precisely when some word meets the bound.
\end{proof}

The same argument applies to exhaustive motion-support BFS for synchronization and goal containment.

\section{Executing and Monitoring Plans}
\label{sec:execution}
A mission supervisor connects the symbolic planner to the robot's motion controller and localizer.
The planner and monitor share the finite transition and output model constructed from the warehouse map.
The supervisor sends the returned word to the controller one action at a time.
The controller reports possible action boundaries and coarse outputs; the localizer supplies pose hypotheses.
The supervisor accepts action completions, maintains the support and output history, and asks the monitor whether a localization update preserves the remaining word's bound.
It then continues execution, requests another word, or stops the robot.
\subsection{Updating the Support}
A plan retains its original support $H$, word $w$, and the model used to compute it.
After prefix $w_i=a_1\cdots a_i$, the motion image is $E_i=\delta(H,w_i)$.
The output trace prefix $\mathbf o_i=o_1\cdots o_i$ gives the more informative support
\begin{equation}
\begin{aligned}
C_i&=H_{w_i,\mathbf o_i}\subseteq E_i,\\
C_0&=H,\qquad C_i=C_{i-1}[a_i,o_i].
\end{aligned}
\label{eq:monitor-support}
\end{equation}
Mapping the localizer's pose hypotheses gives the raw support $L_i$.
The output history selects $C_i$; localizer reports provide additional pose constraints.
Write $R_i$ for the support submitted to the monitor.
The monitor computes $C_i$ from the complete output prefix, then compares it with $R_i$.

If $C_i$ is empty, the trace is incompatible and the monitor returns \outcome{MODEL\_VIOLATION}.
For nonempty $C_i$, it returns \outcome{CONTINUE} when $R_i=C_i$, \outcome{REPLAN} when $\varnothing\ne R_i\subsetneq C_i$, and \outcome{MODEL\_VIOLATION} when $R_i\nsubseteq C_i$.
An empty submitted support $R_i$ also triggers replanning.
All of these checks are meant to ensure a nonempty support before execution continues (and to confirm Assumption~\ref{ass:execution} holds).

The supervisor handles localizer reports in two stages.
First, a nonempty report that is a strict subset of the previously confirmed support must appear again before the supervisor adopts it.
Until then, it keeps the larger support.
After accepting an action completion, it intersects the confirmed report with its current output-updated support.
If the report contains the whole current support, the intersection leaves that support unchanged.
A narrower or partially overlapping report reduces the support and triggers replanning.
Any extra hypotheses in an overlapping report are discarded.
If the supports are disjoint, the supervisor submits the confirmed report unchanged, and the monitor rejects it.

\begin{samepage}
\begin{proposition}[Preserving the remaining bound]
\label{prop:suffix}
Let $w=w_i v$ and let $\varnothing\ne R_i\subseteq C_i$ be an accepted refinement.
For every suffix trace $t$,
\begin{equation}
(R_i)_{v,t}\subseteq H_{w,\mathbf o_i t},\qquad
U_{R_i}(v)\leq U_H(w).
\label{eq:suffix}
\end{equation}
\end{proposition}
\end{samepage}
\begin{proof}
Every state in $R_i$ is the endpoint of some source in $H$ whose prefix trace is $\mathbf o_i$.
A continuation producing $t$ follows the same deterministic transitions from that endpoint.
It therefore belongs to the full-word branch for $\mathbf o_i t$.
Taking cardinalities and then the largest attainable suffix branch proves the bound.
\end{proof}

For synchronization, the weaker inclusion $R_i\subseteq E_i$ already gives $\delta(R_i,v)\subseteq\delta(H,w)$.
For homing or disambiguation, $R_i\subseteq C_i$ is sufficient to inherit the original bound without verifying the suffix again.
Inclusion in $E_i$ alone is insufficient in general.
To see why, consider configurations $p$ and $q$ that remain unchanged under both actions $a$ and $b$.
Action $a$ emits $x$ from $p$ and $y$ from $q$; action $b$ emits $z$ from both.
The word $ab$ homes them.
After observing $x$, filtering only on the later $z$ would incorrectly restore $q$ as a possibility.
The recursive update in Equation~\eqref{eq:monitor-support} retains constraints from the full output history.

The supervisor requests a new word when a localizer refinement strictly reduces the output-conditioned support or the monitor returns \outcome{REPLAN}.
It resets the saved support, output prefix, and completed count together, retaining the accepted localizer refinements.

\subsection{Executing the Actions}
The localizer and coarse classifier differ in scan constraints and may sample at different times.

\textbf{Action boundaries.}
Nominally, \texttt{forward} moves to the next semantic boundary or safely reports blockage.
A left or right turn changes the heading by $90^\circ$ within the current zone.
Cardinal directions henceforth follow the map frame in Figure~\ref{fig:execution}(b), with north up and east right.
The north-facing aisle hypotheses shown there advance to the cross-corridor.
The controller must finish inside the modeled successor region.
Travel parameters may depend on the current support, but must realize the modeled transition $\delta(q,a)$ and output $\eta(q,a)$ for every retained $q$, as required by Assumption~\ref{ass:execution}.

\textbf{Boundary confirmation.}
A candidate is a tentative boundary detection.
The supervisor can accept it, resume the same action, or stop on a model violation.
Resuming leaves the symbolic index and output prefix unchanged.
Acceptance requires one action completion and a corresponding output.
The plan advances only with aligned sensor and localizer evidence and monitor validation.

Before accepting an action completion, the supervisor checks whether the localizer report confirms arrival in the predicted successor region.
A localizer report can confirm the boundary when the candidate's output is predicted from the current support and the entire report lies within the output-conditioned successor support.
A report containing any hypothesis outside the union of the source and successor supports causes a model violation.

\textbf{Observation timing.}
The supervisor requests a localizer report newer than the candidate.
If the robot keeps moving while awaiting acceptance, the scan and report may describe different abstract boundaries.

At each unresolved support, the supervisor first tries synchronization with 1,000 expansions, then homing at $k=1$ with 100,000 expansions after the synchronization budget is exhausted.
The supervisor uses the homing word if one is found; accepted localizer refinements can trigger replanning.
After a singleton is obtained, the supervisor requests a fixed word for goal containment in the target zone and hands over to metric navigation.
Each call has its own expansion budget.

\section{Evaluation}
\label{sec:evaluation}
We evaluate how actions resolve ambiguity during navigation, what fixed sequences cost relative to adaptive plans, and whether sensing and localization satisfy the execution assumptions.
The experiments use a MiR200-sized differential-drive robot in a $36\times56$~m Gazebo warehouse under ROS~2 Jazzy.
Its planar LiDAR provides 360 beams at 10~Hz with a 10~m range.
The model assigns poses to zones and the nearest cardinal heading.
These zones include \emph{bays}, small regions in front of shelf compartments.
The execution model has 319 zones, including the passage through a traversable doorway, and four heading classes per zone, giving 1,276 configurations.
For the solvability comparison in Section~\ref{sec:tradeoffs}, we also consider a model in which that opening is replaced by a wall.
This model omits the doorway zone, leaving 318 zones and 1,272 configurations.
The permitted actions are \texttt{forward}, \texttt{turn\_left}, and \texttt{turn\_right}.

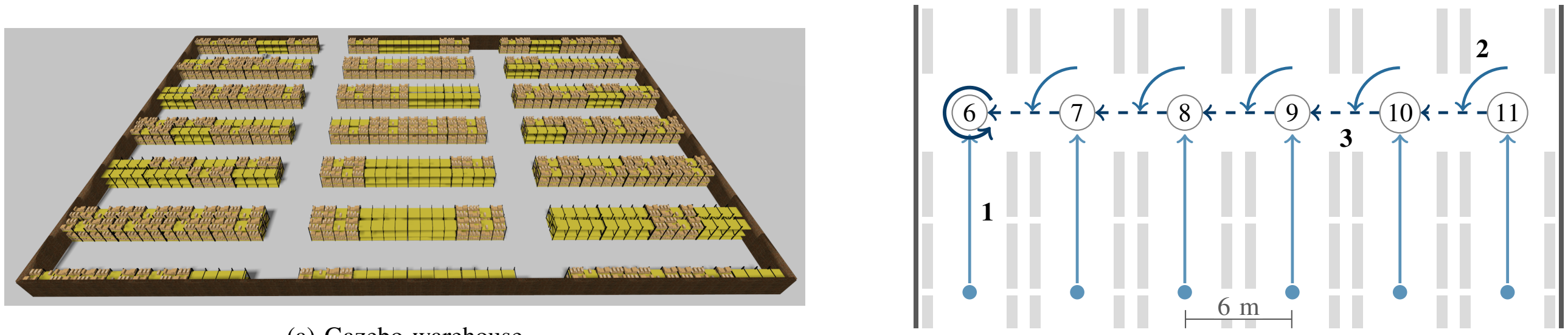
\begin{figure*}[t]
\centering
\begin{minipage}[c]{0.49\textwidth}
\centering
\includegraphics[width=\linewidth]{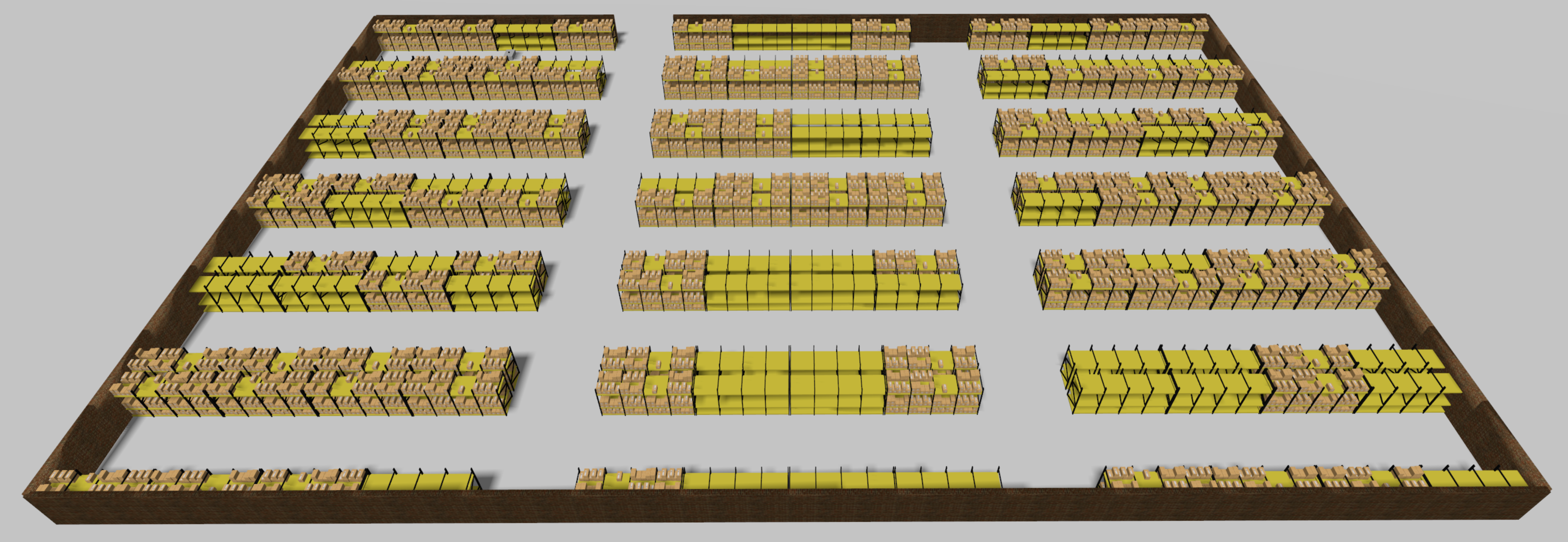}\\
{\footnotesize (a) Gazebo warehouse}
\end{minipage}\hfill
\begin{minipage}[c]{0.49\textwidth}
\centering
\definecolor{stepone}{HTML}{5B93B9}
\definecolor{steptwo}{HTML}{286C9C}
\definecolor{stepthree}{HTML}{083B66}
\begin{tikzpicture}[x=1.95mm,y=1.95mm,every node/.style={font=\footnotesize},
zone/.style={circle,fill=white,draw=black!50,inner sep=1pt,minimum size=3.8mm},
path/.style={->,line width=.9pt}]
\begin{scope}\clip (-.2,26) rectangle (36.2,44);
\fill[black!14] (0.350,24.200) rectangle (0.950,27.800);
\fill[black!14] (0.350,28.200) rectangle (0.950,31.800);
\fill[black!14] (0.350,32.200) rectangle (0.950,35.800);
\fill[black!14] (5.050,24.200) rectangle (5.650,27.800);
\fill[black!14] (5.050,28.200) rectangle (5.650,31.800);
\fill[black!14] (5.050,32.200) rectangle (5.650,35.800);
\fill[black!14] (6.350,24.200) rectangle (6.950,27.800);
\fill[black!14] (6.350,28.200) rectangle (6.950,31.800);
\fill[black!14] (6.350,32.200) rectangle (6.950,35.800);
\fill[black!14] (11.050,24.200) rectangle (11.650,27.800);
\fill[black!14] (11.050,28.200) rectangle (11.650,31.800);
\fill[black!14] (11.050,32.200) rectangle (11.650,35.800);
\fill[black!14] (12.350,24.200) rectangle (12.950,27.800);
\fill[black!14] (12.350,28.200) rectangle (12.950,31.800);
\fill[black!14] (12.350,32.200) rectangle (12.950,35.800);
\fill[black!14] (17.050,24.200) rectangle (17.650,27.800);
\fill[black!14] (17.050,28.200) rectangle (17.650,31.800);
\fill[black!14] (17.050,32.200) rectangle (17.650,35.800);
\fill[black!14] (18.350,24.200) rectangle (18.950,27.800);
\fill[black!14] (18.350,28.200) rectangle (18.950,31.800);
\fill[black!14] (18.350,32.200) rectangle (18.950,35.800);
\fill[black!14] (23.050,24.200) rectangle (23.650,27.800);
\fill[black!14] (23.050,28.200) rectangle (23.650,31.800);
\fill[black!14] (23.050,32.200) rectangle (23.650,35.800);
\fill[black!14] (24.350,24.200) rectangle (24.950,27.800);
\fill[black!14] (24.350,28.200) rectangle (24.950,31.800);
\fill[black!14] (24.350,32.200) rectangle (24.950,35.800);
\fill[black!14] (29.050,24.200) rectangle (29.650,27.800);
\fill[black!14] (29.050,28.200) rectangle (29.650,31.800);
\fill[black!14] (29.050,32.200) rectangle (29.650,35.800);
\fill[black!14] (30.350,24.200) rectangle (30.950,27.800);
\fill[black!14] (30.350,28.200) rectangle (30.950,31.800);
\fill[black!14] (30.350,32.200) rectangle (30.950,35.800);
\fill[black!14] (35.050,24.200) rectangle (35.650,27.800);
\fill[black!14] (35.050,28.200) rectangle (35.650,31.800);
\fill[black!14] (35.050,32.200) rectangle (35.650,35.800);
\fill[black!14] (0.350,40.200) rectangle (0.950,43.800);
\fill[black!14] (5.050,40.200) rectangle (5.650,43.800);
\fill[black!14] (6.350,40.200) rectangle (6.950,43.800);
\fill[black!14] (11.050,40.200) rectangle (11.650,43.800);
\fill[black!14] (12.350,40.200) rectangle (12.950,43.800);
\fill[black!14] (17.050,40.200) rectangle (17.650,43.800);
\fill[black!14] (18.350,40.200) rectangle (18.950,43.800);
\fill[black!14] (23.050,40.200) rectangle (23.650,43.800);
\fill[black!14] (24.350,40.200) rectangle (24.950,43.800);
\fill[black!14] (29.050,40.200) rectangle (29.650,43.800);
\fill[black!14] (30.350,40.200) rectangle (30.950,43.800);
\fill[black!14] (35.050,40.200) rectangle (35.650,43.800);
\fill[black!65] (-0.125,-0.125) rectangle (0.125,56.125);
\fill[black!65] (35.875,-0.125) rectangle (36.125,56.125);
\end{scope}
\fill[stepone] (3.0,28.0) circle (.4);
\draw[path,stepone] (3.0,28.5) -- (3.0,36.8);
\node[zone] (z0) at (3.0,38.0) {6};
\fill[stepone] (9.0,28.0) circle (.4);
\draw[path,stepone] (9.0,28.5) -- (9.0,36.8);
\node[zone] (z1) at (9.0,38.0) {7};
\draw[path,steptwo] (9.0,40.5) arc[start angle=90,end angle=178,radius=2.5];
\fill[stepone] (15.0,28.0) circle (.4);
\draw[path,stepone] (15.0,28.5) -- (15.0,36.8);
\node[zone] (z2) at (15.0,38.0) {8};
\draw[path,steptwo] (15.0,40.5) arc[start angle=90,end angle=178,radius=2.5];
\fill[stepone] (21.0,28.0) circle (.4);
\draw[path,stepone] (21.0,28.5) -- (21.0,36.8);
\node[zone] (z3) at (21.0,38.0) {9};
\draw[path,steptwo] (21.0,40.5) arc[start angle=90,end angle=178,radius=2.5];
\fill[stepone] (27.0,28.0) circle (.4);
\draw[path,stepone] (27.0,28.5) -- (27.0,36.8);
\node[zone] (z4) at (27.0,38.0) {10};
\draw[path,steptwo] (27.0,40.5) arc[start angle=90,end angle=178,radius=2.5];
\fill[stepone] (33.0,28.0) circle (.4);
\draw[path,stepone] (33.0,28.5) -- (33.0,36.8);
\node[zone] (z5) at (33.0,38.0) {11};
\draw[path,steptwo] (33.0,40.5) arc[start angle=90,end angle=178,radius=2.5];
\draw[path,stepthree,dashed] (z1.west) -- (z0.east);
\draw[path,stepthree,dashed] (z2.west) -- (z1.east);
\draw[path,stepthree,dashed] (z3.west) -- (z2.east);
\draw[path,stepthree,dashed] (z4.west) -- (z3.east);
\draw[path,stepthree,dashed] (z5.west) -- (z4.east);
\draw[path,stepthree,line width=1.1pt] (3,38) ++(30:1.4) arc[start angle=30,end angle=325,radius=1.4];
\node[black,font=\footnotesize\bfseries,fill=white,inner sep=1pt] at (4,32.5) {1};
\node[black,font=\footnotesize\bfseries,fill=white,inner sep=1pt] at (31.6,41.6) {2};
\node[black,font=\footnotesize\bfseries,fill=white,inner sep=1pt] at (24,36.6) {3};
\draw[|-|,black!65] (15,26.5) -- (21,26.5) node[midway,above,fill=white,inner sep=1pt] {6 m};
\end{tikzpicture}\\
{\footnotesize (b) Map detail: $6\to5$ by motion, $5\to1$ by blockage} %
\end{minipage}
\caption{The Gazebo screenshot (a) shows one of our test warehouses; (b) shows a detail of its abstract map, illustrating how a blocked path resolves six aisle hypotheses without localizer updates.
The map shows modeled transitions between zone centers and defines the cardinal directions: north is up, and east is right.
The action sequence is (1) forward, (2) turn left, (3) forward, shown in increasing blue darkness.
Dots mark possible starts and circles identify intersection zones; curved arrows show turns in place.
The last move merges six possibilities into five endpoints.
Observed blockage identifies intersection 6 facing west (dark self-loop); a junction output leaves five alternatives along the dashed moves.
}
\label{fig:execution}
\end{figure*}

Figure~\ref{fig:execution} shows the warehouse geometry and an execution prefix that resolves six aisle hypotheses when the robot encounters a blocked path.
We analyze eleven navigation trials and compare planners on the warehouse model with a traversable doorway.
The Appendix gives the measurement and replay procedures.

\subsection{Sources of Ambiguity Reduction}
\label{sec:attribution}
Figure~\ref{fig:execution} shows the first three actions of a selected seven-action synchronizing word.
The first forward action and left turn both report a junction from every hypothesis.
On the next forward action, the westernmost hypothesis stays at the wall and reports blockage. Its neighbor reaches the same intersection, and all five moving hypotheses report a junction.
Motion therefore reduces six configurations to five endpoints.
Blockage identifies one configuration, whereas a junction output leaves five, giving the displayed prefix $u$ worst-case ambiguity $U_H(u)=5$.
After observing blockage, the supervisor requests a route to the goal.
We count support decreases successively after motion merging, output filtering and localizer filtering, using Equation~\eqref{eq:posterior} and Section~\ref{sec:execution}.
Five runs start with ten hypotheses and continue to accept localizer updates.
Six runs start with supplied supports: two from six aisle hypotheses and four from a symmetric pair, with localizer updates disabled.
Table~\ref{tab:attribution} reports the total support decreases for each mechanism over the actions in each group.

\begin{table}[t]
\centering
\caption{Sequential support reductions during navigation.}
\label{tab:attribution}
\footnotesize
\setlength{\tabcolsep}{4pt}
\begin{tabular}{@{}lrrrr@{}}
\toprule
& & \multicolumn{3}{c}{Total support decreases}\\
\cmidrule(l){3-5}
Support updates & Runs & Merging & Coarse outputs & Localizer\\
\midrule
Live localizer & 5 & 0 & 0 & 45\\
Supplied support & 6 & 2 & 12 & Disabled\\
\bottomrule
\end{tabular}
\end{table}

All reductions in the live-localizer runs come from localizer filtering.
Coarse outputs and merging contribute to the supplied-support runs.

For each trial, we take the sequence of executed actions and observed outputs from its log.
We replay this sequence in the model without localizer refinements.
This leaves eight hypotheses in every live-localizer run and one in every supplied-support run.

Ground-truth replay finds the true configuration in all eleven initial supports and after all 43 ambiguity-resolution actions.
Every run reaches its requested goal region.

\subsection{Solvability and Planning Cost}
\label{sec:tradeoffs}
\textbf{Doorway geometry.}
Two configurations are behaviorally equivalent when every common word produces the same output trace from them.
Table~\ref{tab:classes} compares the wall and traversable-doorway models under the same three actions and output names.

\begin{table}[t]
\centering
\caption{Equivalence classes by size and doorway model.}
\label{tab:classes}
\footnotesize
\begin{tabular}{@{}lrrrrr@{}}
\toprule
Model & $|Q|$ & Classes & Size 1 & Size 2 & Size 4\\
\midrule
Wall & 1272 & 540 & 0 & 444 & 96\\
Doorway & 1276 & 1084 & 892 & 192 & 0\\
\bottomrule
\end{tabular}
\end{table}

Exhaustive search finds 828 unmergeable pairs among the wall model's $444+6\cdot96=1{,}020$ equivalent unordered pairs.
A support containing any such pair cannot be homed.

The doorway model has 192 size-two equivalence classes (Table~\ref{tab:classes}).
They pair corresponding bays on opposite aisle sides with the same heading: 48 bay-zone pairs across four headings.
Each pair merges at a common intersection and heading within five actions.
The common endpoint depends on the pair.
For example, \texttt{forward} merges two south-facing bay hypotheses at intersection 0.
The starting bay remains unknown, but the endpoint is known.

Thus every pair in the doorway model can be merged or distinguished using the permitted actions.
Appending witnesses for unresolved pairs yields a fixed homing word for every support~\cite{Cirisci2019Homing}.

We compare the pair-proposal planner with exact fixed-word search, iterative pair homing (Greedy-HS~\cite{Cirisci2019Homing}), and adaptive planning on the traversable-doorway model, using the same supports, actions and unit action costs.
The suite contains 158 requests: eighteen recorded homing requests over four supports and 140 sampled requests.
We sample twenty supports at each size in $\{2,3,4,6,8,12,16\}$.
Sampled supports also include bay hypotheses.
Sampled goal requests target intersection 0, outside Figure~\ref{fig:execution}(b); recorded requests retain their goal zones.
Any heading is allowed.

Fixed searches have a budget of $10^4$ expansions per request.
The adaptive baseline computes minimax reachability over the reachable supports, choosing an action after each output~\cite{Bonet2009DetPOMDP}.
Its terminal condition is $|S|\leq1$ for homing and $S\subseteq T$ for goal containment.
Every reachable-support graph closes below its 50,000-support limit, and every returned policy terminates on all output branches.
We compare planning success and the worst-case action count of each returned plan.
In Table~\ref{tab:planners}, Search uses exact homing search with or without pair proposals; homing-plus-route appends goal routes to the returned words.

\begin{table}[t]
\centering
\caption{Planning success on twenty sampled supports per size.}
\label{tab:planners}
\footnotesize
\setlength{\tabcolsep}{4pt}
\begin{tabular}{@{}rrrrrrrr@{}}
\toprule
& & \multicolumn{3}{c}{Homing} & \multicolumn{3}{c}{Goal}\\
\cmidrule(lr){3-5}\cmidrule(l){6-8}
$|H|$ & Sync & Search & \shortstack{Greedy\\HS} & Adaptive & \shortstack{Fixed\\word} & \shortstack{Homing\\+ route} & Adaptive\\
\midrule
2 & 5 & 20 & 20 & 20 & 20 & 20 & 20\\
3 & 6 & 20 & 20 & 20 & 4 & 20 & 20\\
4 & 0 & 20 & 20 & 20 & 0 & 20 & 20\\
6 & 0 & 20 & 20 & 20 & 0 & 20 & 20\\
8 & 0 & 18 & 20 & 20 & 0 & 18 & 20\\
12 & 0 & 13 & 20 & 20 & 0 & 13 & 20\\
16 & 0 & 3 & 20 & 20 & 0 & 3 & 20\\
\bottomrule
\end{tabular}
\end{table}

\textbf{Adaptation.}
Adaptive homing solves all 158 requests.
On the recorded supports its worst-case action count equals that of both homing searches.
On 52 of the 114 jointly solved sampled supports it uses fewer actions; the adaptive-to-fixed ratio ranges from 0.54 to 1, with median 1.
Table~\ref{tab:planners} shows that bounded fixed-word search increasingly reaches its budget as support size grows.
Its synchronization column tests the stronger objective that all endpoints coincide, regardless of observations.
For synchronization, fifteen size-two supports have no solution by exhaustive search; every other unsuccessful fixed search in the table ends at the expansion limit.

We fix each word returned by the two homing searches, then append shortest goal routes from its endpoints.
This composition matches adaptive goal planning's worst-case action count on the recorded supports.
On the 114 sampled supports with a search solution, the adaptive-to-composed cost ratio ranges from 0.50 to 1, with a median of 0.83.
Routing at the first singleton posterior, as in Figure~\ref{fig:execution}, reduces the worst-case cost on 48 of these supports and raises the median ratio to 0.89.

Fixed-word goal planning, which uses one word for every hypothesis, solves six recorded and 24 sampled requests.
Adaptive goal planning solves every request and uses fewer actions on all thirty jointly solved requests, with median ratios of 0.59 and 0.58, respectively.

\textbf{Pair construction.}
The exact-only variant suppresses pair proposals while retaining the same family BFS and final verification.
Both searches solve all eighteen recorded requests and 114 sampled requests, with equal word lengths; both exhaust their budget on the other 26.
Restricting the pair table itself to the three permitted actions leaves these outcomes and lengths unchanged.
Greedy-HS repeatedly appends a shortest witness for an unresolved pair and updates all remaining pairs.
It homes all 158 supports, including the 26 requests where search reaches its budget.
On the 114 sampled supports solved by both approaches, Greedy-HS words have a median length ratio of 1.22 relative to search, with ratios from 1 to 2.33.
Thus, iterative pair construction resolves the larger supports, while exact search returns shorter words on many shared instances.

\subsection{Observation Agreement, Model Sensitivity \& Limitations}
\label{sec:validity}
All eleven navigation trials reach their goals.
We next vary poses and headings to examine when the output agreement required by Assumption~\ref{ass:execution} holds and how disagreements affect planning.
The modeled classifier counts graph-neighbor bits; the live classifier tests four body-relative $20^\circ$ scan sectors.
A sector is open when at least $60\%$ of its beams reach 4~m, and both classifiers require three open directions for a junction.
Heading and position can change the scan-sector count while the graph-neighbor count stays fixed.

\textbf{Heading and action regions.}
The heading study selects 31 aisle and corridor rectangles and all four cardinal headings, giving 124 zone-heading entries.
Map-cast scans cover nine positions per rectangle and offsets $0^\circ$, $\pm1^\circ$, $\pm2^\circ$, $\pm4.5^\circ$, and $\pm9^\circ$.
An entry agrees only when every sampled position and heading produces its modeled output.
Table~\ref{tab:heading} accumulates both signs through each stated magnitude.
All 124 entries agree at cardinal headings.
Each has at least one disagreement among the sampled offsets through $\pm9^\circ$, near the largest recorded stopping error.

\begin{table}[t]
\centering
\caption{Model agreement across sampled heading offsets.}
\label{tab:heading}
\footnotesize
\begin{tabular}{@{}lrrrr@{}}
\toprule
Largest tested offset ($^\circ$) & 0 & $\pm1$ & $\pm2$ & $\pm9$\\
\midrule
Matching zone--heading entries / 124 & 124 & 122 & 72 & 0\\
\bottomrule
\end{tabular}
\end{table}

Actions end in more specific regions than the full drawn rectangles.
The execution model extends intersection regions into unnamed corridor gaps.
The endpoint study samples landing bands within these regions, including the turn positions reached from them.
Each entry is a state--action pair and agrees only when every sampled landing pose and heading gives its modeled output.
With a 0.45~m boundary inset and sideways displacement limited to $\pm1.04$~m, 2,997 of 3,198 scan-dependent state--action entries agree through the sampled $\pm9^\circ$ offsets.
Across the full inset width, 2,754 agree.
The narrower band was selected from the 85 recorded endpoints; 201 entries contain at least one output disagreement.

\textbf{Planning impact.}
To measure the effect on homing, we remove entries with sampled output disagreements from the planning model.
Each exclusion covers the entire region, including poses with agreeing outputs.
Scan-free blocked self-loops remain available.
An action is allowed only when its entry remains for every state in the current support.
Under this restriction, adaptive homing solves 66 of 158 requests; exhaustive search finds no policy for the other 92, including all four recorded supports.

\textbf{Wall-turn checks.}
After a wall stop at intersection 11, the model predicts a junction on turning, whereas the map-cast output is passage at a cardinal heading.
The same output disagreement occurs at two other intersections.
Scripted forward--left-turn commands using supplied supports reproduce it at all three wall locations.
Separate left-turn tests from the twelve witness poses also disagree with the model.
These checks identify wall-turn poses where Assumption~\ref{ass:execution}'s output agreement is not met.

\textbf{Boundary checks.}
Two further trials record scans and localizer reports at action boundaries.
Every output is reproduced from its selected scan or blockage rule.
The supplied-support run reaches its goal.
The live-localizer run stops before monitor validation because its boundary report includes unrelated configurations alongside the true successor.

\textbf{Initial supports.}
A separate study restarts localization at fixed poses and checks the first fresh report.
Of 512 held-out trials, 29 stop at pose reset.
The remaining 483 produce a first localization report: 453 include the true zone and heading and 30 exclude it.
Exclusions occur at some bay-strip, intersection, and threshold poses.

\section{Conclusion}
\label{sec:conclusion}
In this work, we proposed a framework for planning to reach a goal or reduce uncertainty: Verified fixed words bound residual ambiguity for every output trace in the finite model.
A monitor preserves the remaining bound under Proposition~\ref{prop:suffix}'s support refinements.
Supplied-support runs resolve ambiguity through merging and coarse observations before reaching their goals. %
In our experiments, homing search followed by goal routing matches adaptive planning's worst-case action count.
Iterative pair homing solves the requests where bounded search stops, at the cost of longer words on many shared instances.
Routing at the first singleton reduces the worst-case action count to the goal on sampled supports.
The additional checks identify where wall-turn observations disagree with the model and initial reports omit the true configuration.

The architecture is promising. Our results demonstrate it
does manage to resolve ambiguity and reaches goals in simulation.
We expect it to support sonar or camera localization with suitable classifiers and action/output models.

\section*{Acknowledgment}
OpenAI Codex assisted with writing and experiments.

\bibliographystyle{IEEEtran}
\bibliography{refs}

\appendix[Experimental Procedures]
\label{app:methods}
\textbf{Planning comparisons.}
Adaptive values are computed backward from terminal supports, with each selected policy edge decreasing the value.
Greedy-HS chooses a shortest resolving witness among the current unresolved pairs, breaking ties by pair index, and discards merged or distinguished pairs after each appended witness.
Its pair table uses the three permitted actions.
All returned words and policy branches are independently replayed.
Early goal routing follows each fixed word until its first singleton posterior, then takes a shortest path to the goal; costs maximize over all initial hypotheses.

\textbf{Initialization.}
The initialization study uses 128 discovery poses and 512 held-out poses, equally allocated to sixteen strata: aisle, intersection, threshold and bay regions crossed with four cardinal headings.
Positions are proposed uniformly by area and heading errors uniformly within $\pm9^\circ$, rejecting chassis--obstacle intersections.
Localization restarts at each pose with unchanged parameters and initial domain.
We score the first complete fresh report after settling, allowing 30 wall-clock seconds after launch.
Each published hypothesis centroid is assigned a map zone and the nearest cardinal heading; coverage requires the true zone and heading in that support.
Bay strips map to aisle states.
Availability and coverage are fractions of all assigned poses.

\textbf{Execution checks.}
The navigation trials are checked by independently replaying their plans and support updates against the finite model.
For the two boundary tests, we reconstruct the observations and support filtering from the recorded scans and localizer reports.
Wall approaches use supplied intersection hypotheses $\{(5,\mathrm{east}),(11,\mathrm{east}),(6,\mathrm{west}),(0,\mathrm{east})\}$.
The forward action blocks and removes intersection 0; each survivor then predicts a junction on the left turn, whereas the robot reads passage.

\textbf{Fault response.}
Six injected faults cover missing/duplicate outputs, a corrupted prefix, database refusal, and missing/duplicate responses, each repeated three times while coasting.
All eighteen trials cancel the active motion command, brake on the next controller update, and reach physical rest within five simulated seconds.
Rest requires sampled speeds below 0.01~m/s and 0.01~rad/s for a nominal 0.3~s.
\label{page:main-last}
\end{document}